\documentclass[letterpaper, 10pt,twocolumn]{article}
\pdfoutput=1

\newcommand{\adstitle}{Multi-camera Realtime 3D Tracking of Multiple Flying Animals}

\date{}

\usepackage{fancyhdr}

\usepackage{ifpdf}
\ifpdf
\usepackage[%
  pdftitle={\adstitle},
  pdfauthor={Andrew D. Straw, Kristin Branson, Titus R. Neumann and Michael H. Dickinson},%
  bookmarks=true,%
  bookmarksopen=true,%
  breaklinks=true,%
  colorlinks=true,%
  linkcolor=black,anchorcolor=black,%
  citecolor=black,filecolor=black,%
  menucolor=black,pagecolor=black,%
  urlcolor=black]{hyperref}
\else
\usepackage[%
  breaklinks=true,%
  colorlinks=true,%
  linkcolor=black,anchorcolor=black,%
  citecolor=black,filecolor=black,%
  menucolor=black,pagecolor=black,%
  urlcolor=black]{hyperref}
\fi

\usepackage{graphics}
\usepackage{epsfig}
\usepackage{amsmath}
\usepackage{epsfig, latexsym, amsmath, amsfonts, amssymb,graphicx}
\usepackage{amsmath,amsthm,amssymb}
\usepackage{graphicx, epsfig}
\usepackage{natbib}
\usepackage{url}

\DeclareMathOperator*{\argmax}{argmax}

\newcommand{\cross}{\times}
\newcommand{\trnsps}{\mathsf{T}}
\newcommand{\eqnref}[1]{\ref{#1}}

\newcommand{\fig}{Figure }
\newcommand{\figs}{Figures }

\newcommand{\numcams}{n}
\newcommand{\twodeevec}{\boldsymbol{{\mathrm x}}}
\newcommand{\threedeevec}{\boldsymbol{{\mathrm X}}}
\newcommand{\cammatrix}{{\mathrm P}}

\newcommand{\dehomogenize}{{\mathcal H}}

\newcommand{\linefunc}{{\mathcal L}}

\newcommand{\onecolfig}{9cm}
\newcommand{\onefivecolfig}{14cm}

\newcommand{\genericvector}{\boldsymbol{{\mathrm{v}}}}
\newcommand{\genericmatrix}{{\mathrm{M}}}

\newcommand{\kalmanA}{{\mathrm A}} 
\newcommand{\allstate}{\mathcal{S}}
\newcommand{\kstatevec}{\boldsymbol{\mathrm{s}}}
\newcommand{\kobsvec}{\boldsymbol{\mathrm{y}}}
\newcommand{\kalmanv}{\boldsymbol{\mathrm{v}}} 
\newcommand{\kalmanw}{\boldsymbol{\mathrm{w}}} 
\newcommand{\kalmanR}{{\mathrm R}} 
\newcommand{\kalmanQ}{{\mathrm Q}} 
\newcommand{\kalmanC}{{\mathrm H}_t} 
\newcommand{\kalmanh}{h} 

\newcommand{\kesterr}{\mathrm{P}}
\newcommand{\kstateest}  {\hat{\kstatevec}}
\newcommand{\kstateapriori}  {\hat{\kstatevec}_{t \vert t-1}}
\newcommand{\kstateapost}    {\hat{\kstatevec}_{t \vert t}}
\newcommand{\kstateapostprev}{\hat{\kstatevec}_{t-1 \vert t-1}}
\newcommand{\kesterrapriori}  {\kesterr_{t \vert t-1}}
\newcommand{\kesterrapost}    {\kesterr_{t \vert t}}
\newcommand{\kesterrapostprev}{\kesterr_{t-1 \vert t-1 }}

\newcommand{\kalmangaintimeless}{{\mathrm K}}
\newcommand{\kalmangain}{{\kalmangaintimeless_t}}
\newcommand{\eyemat}{{\mathrm I}}

\newcommand{\Er}{\bar{r}}
\newcommand{\Es}{\bar{s}}
\newcommand{\Et}{\bar{t}}
\newcommand{\Eu}{\bar{u}}
\newcommand{\Ev}{\bar{v}}

\newcommand{\numtwodpoints}{m}
\newcommand{\zee}{\boldsymbol{\mathrm{z}}}
\newcommand{\Zee}{\mathrm{Z}}
\newcommand{\allZee}{\mathcal{Z}}
\newcommand{\assignvec}{\boldsymbol{d}}
\newcommand{\dassocvar}{\mathcal{D}}

\newcommand{\veca}{\boldsymbol{\mathrm{a}}}
\newcommand{\vecabar}{\boldsymbol{\mathrm{\bar{a}}}}

\newcommand{\indicator}{\boldsymbol{\mathrm{1}}}

\title{\adstitle}

\author{Andrew D. Straw, Kristin Branson, Titus R. Neumann \& Michael H. Dickinson \\
California Institute of Technology \\
Bioengineering, Mailcode 138-78 \\
Pasadena, CA 91125 USA}

\begin{document}

\maketitle

\section*{Abstract}

Automated tracking of animal movement allows analyses that would not
otherwise be possible by providing great quantities of data. The
additional capability of tracking in realtime---with minimal
latency---opens up the experimental possibility of manipulating
sensory feedback, thus allowing detailed explorations of the neural
basis for control of behavior. Here we describe a new system capable
of tracking the position and body orientation of animals such as flies
and birds. The system operates with less than 40 msec latency and can
track multiple animals simultaneously. To achieve these results, a
multi target tracking algorithm was developed based on the Extended
Kalman Filter and the Nearest Neighbor Standard Filter data
association algorithm. In one implementation, an eleven camera system
is capable of tracking three flies simultaneously at 60 frames per
second using a gigabit network of nine standard Intel Pentium 4 and
Core 2 Duo computers.  This manuscript presents the rationale and
details of the algorithms employed and shows three implementations of
the system. An experiment was performed using the tracking system to
measure the effect of visual contrast on the flight speed of {\em
  Drosophila melanogaster}. At low contrasts, speed is more variable
and faster on average than at high contrasts. Thus, the system is
already a useful tool to study the neurobiology and behavior of freely
flying animals. If combined with other techniques, such as `virtual
reality'-type computer graphics or genetic manipulation, the tracking
system would offer a powerful new way to investigate the biology of
flying animals.

Keywords: multitarget tracking; Kalman filter; computer vision; data
association; Drosophila; insects; birds; hummingbirds; neurobiology;
animal behavior

\section{Introduction}\label{introduction}

Much of what we know about the visual guidance of flight
\citep{Schilstra_1999, Kern_2005, Srinivasan_1996, Tammero_2002},
aerial pursuit \citep{Land_Collett_1974, buelthoff_1980,
  wehrhahn_1982, Wagner_1986, Mizutani_2003}, olfactory search
algorithms \citep{Frye_2003, Budick_2006}, and control of aerodynamic
force generation \citep{David_1978, Fry_2003} is based on experiments
in which an insect was tracked during flight.  To facilitate these
types of studies and to enable new ones, we created a new, automated
animal tracking system. A significant motivation was to create a
system capable of robustly gathering large quantities of accurate data
in a highly automated fashion in a flexible way. The realtime nature
of the system enables experiments in which an animal's own movement is
used to control the physical environment, allowing virtual-reality or
other dynamic stimulus regimes to investigate the feedback-based
control performed by the nervous system. Furthermore, the ability to
easily collect flight trajectories facilitate data analysis and
behavioral modeling using machine learning approaches that require
large amounts of data.

Our primary innovation is the use of arbitrary numbers of inexpensive
cameras for markerless, realtime tracking of multiple targets.
Typically, cameras with relatively high temporal resolution, such as
100 frames per second, and which are suitable for realtime image
analysis (those that do not buffer their images to on-camera memory),
have relatively low spatial resolution. To have high spatial
resolution over a large tracking volume, many cameras are
required. Therefore, the use of multiple cameras enables tracking over
large, behaviorally- and ecologically- relevant spatial scales with
high spatial and temporal resolution while minimizing the effects of
occlusion. The use of multiple cameras also gives the system its name,
{\em flydra}, from `fly', our primary experimental animal, and the
mythical Greek multi-headed serpent `hydra'.

Flydra is largely composed of standard algorithms, hardware, and
software. Our effort has been to integrate these disparate pieces of
technology into one coherent, working system with the important
property that the multi-target tracking algorithm operates with low
latency during experiments.

\subsection{System overview}\label{algo_overview}

A Bayesian framework provides a natural formalism to describe our
multi-target tracking approach. In such a framework, previously held
beliefs are called the {\em a priori}, or prior, probability
distribution of the state of the system. Incoming observations are
used to update the estimate of the state into the {\em a posteriori},
or posterior, probability distribution. This process is often likened
to human reasoning, whereby a person's best guess at some value is
arrived at through a process of combining previous expectations of
that value with new observations that inform about the value.

The task of flydra is to find the maximum {\em a posteriori} (MAP)
estimate of the state $\allstate_t$ of all targets at time $t$ given
observations $\allZee_{1:t}$ from all time steps (starting with the
first time step to the current time step), $p(\allstate_t \vert
\allZee_{1:t} )$.  Here, $\allstate_t$ represents the state (position
and velocity) of all targets, $\allstate_t =
(\kstatevec^1_{t},\ldots,\kstatevec^{l_t}_{t})$ where $l_t$ is the number
of targets at time $t$. Under the first-order Markov assumption, we
can factorize the posterior as
\begin{equation} \label{eqn:simple_bayes}
p(\allstate_t \vert \allZee_{1:t} ) \propto
p(\allZee_{t} \vert \allstate_t )
\int p(\allstate_t \vert \allstate_{t-1}) p(\allstate_{t-1} \vert \allZee_{1:t-1})
d\allstate_{t-1} .
\end{equation} Thus, the process of estimating the posterior probability of
target state at time $t$ is a recursive process in which new
observations are used in the model of observation likelihood
$p(\allZee_{t} \vert \allstate_t )$. Past observations become
incorporated into the prior, which combines the motion model
$p(\allstate_t \vert \allstate_{t-1})$ with the target probability
from the previous time step $p(\allstate_{t-1} \vert
\allZee_{1:t-1})$.

Flydra utilizes an Extended Kalman Filter to approximate the solution
to equation \eqnref{eqn:simple_bayes}, as described in Section
\ref{kalman_filtering}. The observation $\allZee_{t}$ for each time
step is the set of all individual low-dimensional feature vectors
containing image position information arising from the camera views of
the targets (Section \ref{2D_features}). In fact, equation
\eqnref{eqn:simple_bayes} neglects the challenges of data association
(linking individual observations with specific targets) and targets
entering and leaving the tracking volume. Therefore the Nearest
Neighbor Standard Filter data association step is used to link
individual observations with target models in the model of observation
likelihood (Section \ref{data_association}), and the state update
model incorporates the ability for targets to enter and leave the
tracking volume (Section \ref{birth_model}). The heuristics employed
to implement the system typically were optimizations with regard to
realtime performance and low latency rather than a compact form, and
our system only approximates the full Bayesian solution rather than
perfectly implements it.  Nevertheless, the remaining sections of this
manuscript address their relation to the global Bayesian framework
where possible. Aspects of the system which were found to be important
for low-latency operation are mentioned.

\begin{figure*}
\includegraphics[width=\onefivecolfig]{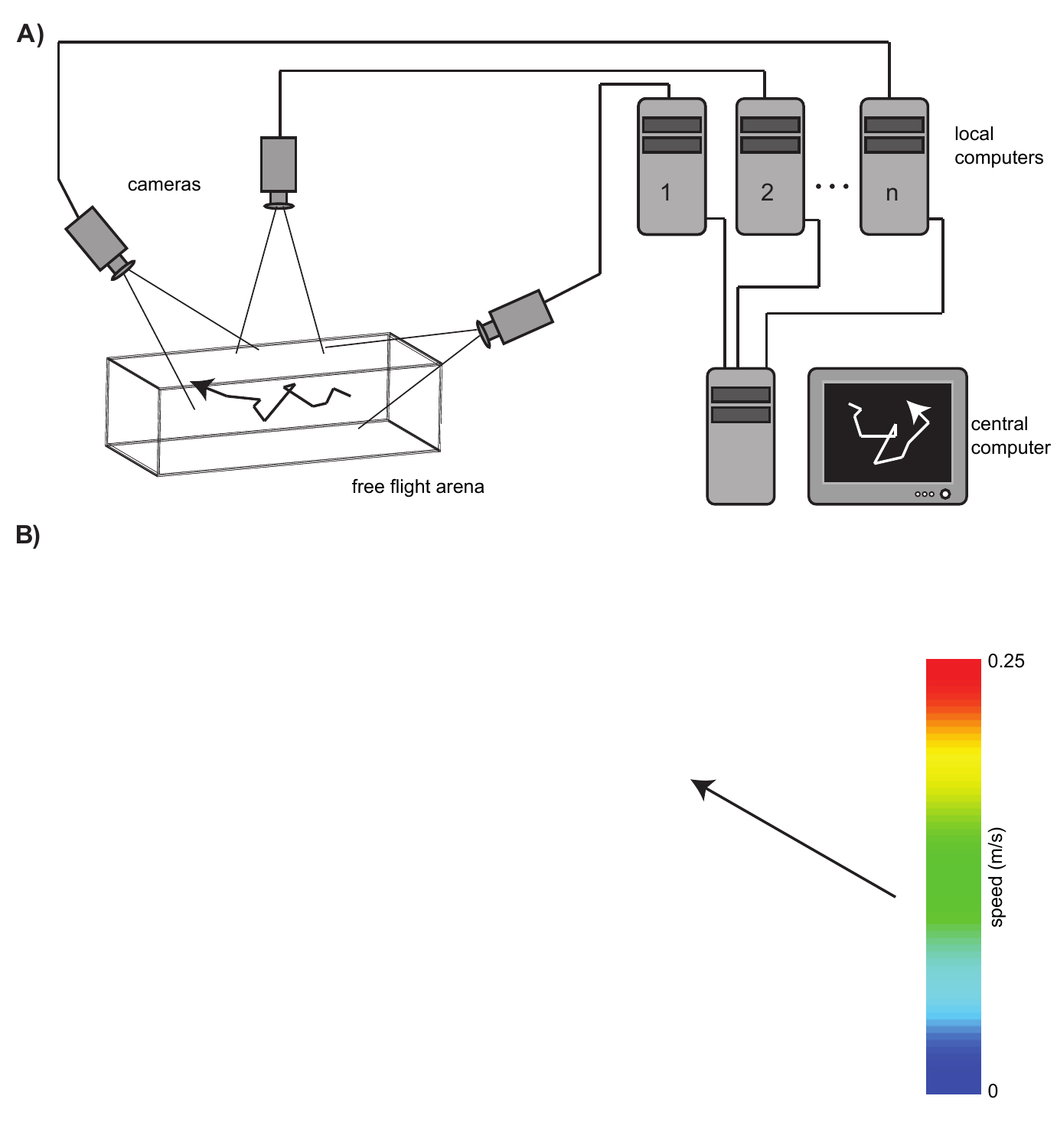}

\caption{A) Schematic diagram of the multi-camera tracking system. B)
  A trajectory of a fly ({\em Drosophila melanogaster}) near a dark,
  vertical post. Arrow indicates direction of flight at onset of
  tracking.}
\label{fig:overview}
\end{figure*}

\begin{figure*}
\includegraphics[width=\onefivecolfig]{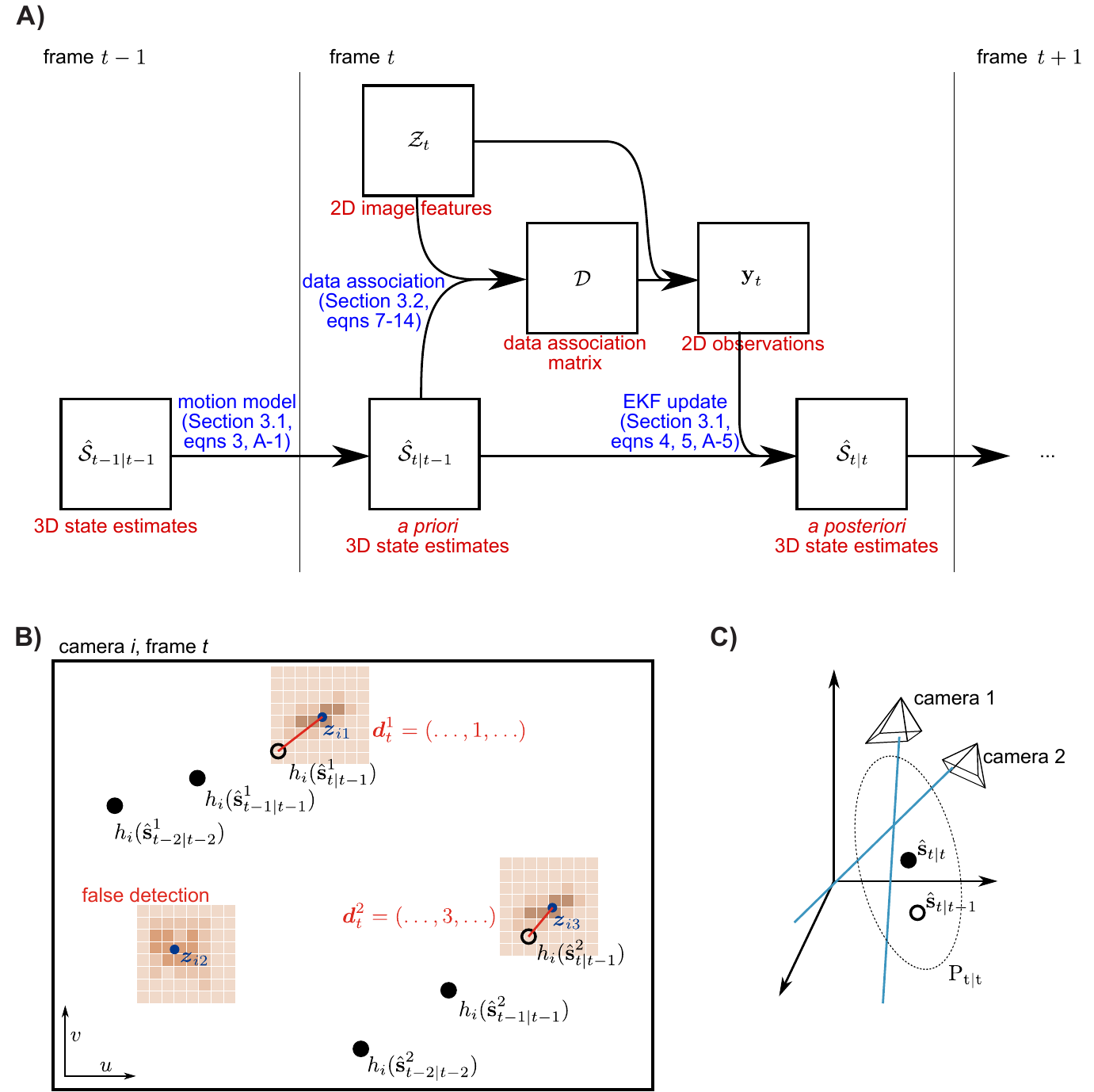}

\caption{A) Flowchart of operations. B) Schematic illustration of a 2D
  camera view showing the raw images (brown), feature extraction
  (blue), state estimation (black), and data association (red). C) 3D
  reconstruction using the Extended Kalman filter uses prior state
  estimates (open circle) and observations (blue lines) to construct a
  posterior state estimate (closed circle) and covariance ellipsoid
  (dotted ellipse).}
\label{fig:flowchart}
\end{figure*}

\begin{figure*}
\includegraphics[width=\onefivecolfig]{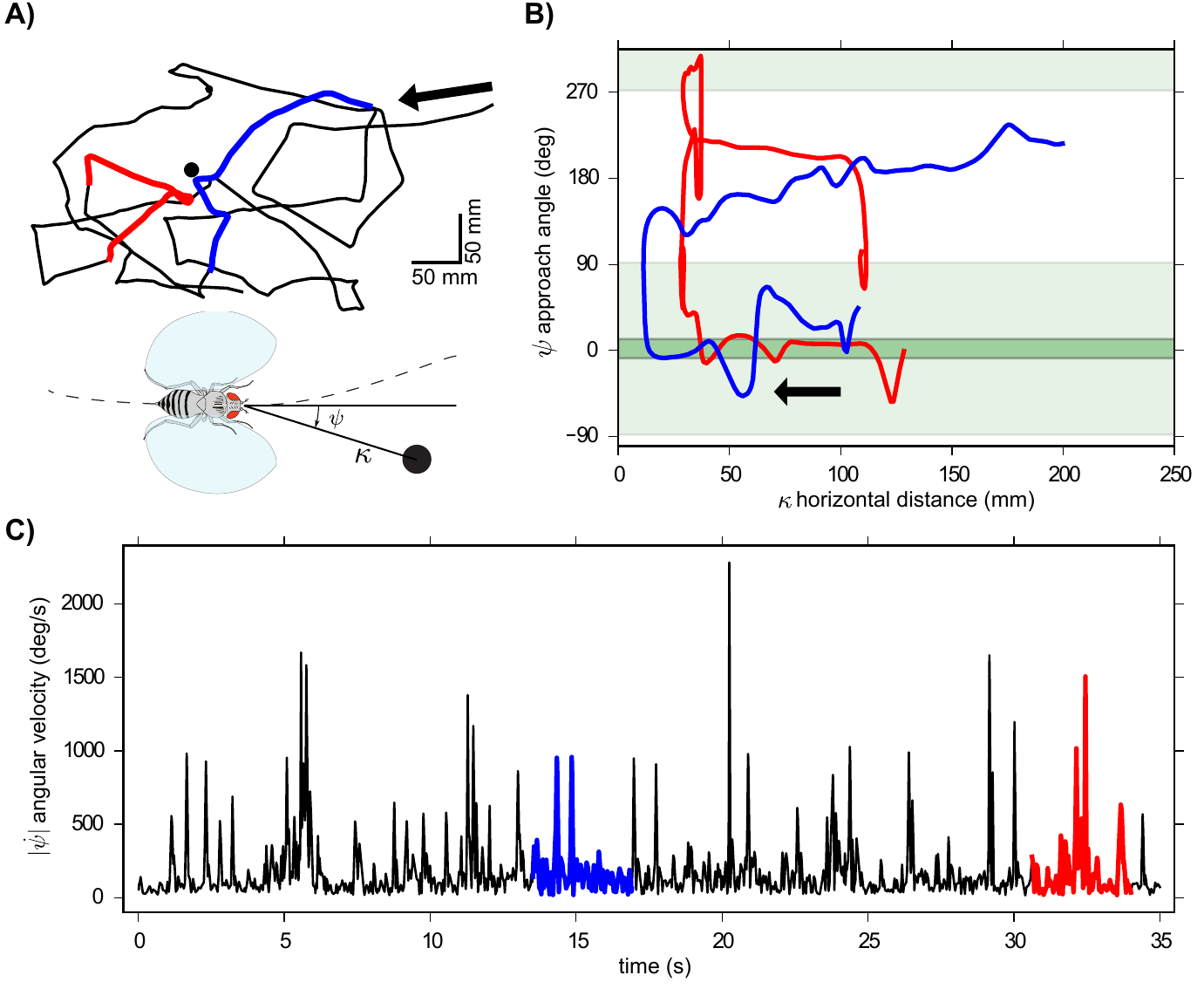}

\caption{A) Top view of the fly trajectory in \fig
  \ref{fig:overview}B, showing several close approaches to and
  movements away from a dark post placed in the center of the
  arena. The arrow indicates initial flight direction. Two sequences
  are highlighted in color. The inset shows the coordinate system. B)
  The time-course of attraction and repulsion to the post is
  characterized by flight directly towards the post until only a small
  distance remains, at which point the fly turns rapidly and flies
  away. Approach angle $\psi$ is quantified as the difference between
  the direction of flight and bearing to the post, both measured in
  the horizontal plane. The arrow indicates the initial approach for
  the selected sequences. C) Angular velocity (measured tangent to the
  3D flight trajectory) indicates that relatively straight flight is
  punctuated by saccades (brief, rapid turns).  }

\label{fig:analysis}
\end{figure*}

The general form of the apparatus is illustrated in \fig
\ref{fig:overview}A and a flowchart of operations is given in \fig
\ref{fig:flowchart}A.  Digital cameras are connected (with an IEEE
1394 FireWire bus or a dedicated gigabit ethernet cable) to image
processing computers that perform a background subtraction based
algorithm to extract image features such as the 2D target position and
orientation in a given camera's image. From these computers, this 2D
information is transmitted over a gigabit ethernet LAN to a central
computer, which performs 2D-to-3D triangulation and tracking.
Although the tracking results are generated and saved online, in
realtime as the experiment is performed, raw image sequences can also
be saved for both verification purposes as well as other types of
analyses. Finally, reconstructed flight trajectories, such as that of
\fig \ref{fig:overview}B, may then be subjected to further analysis,
as in \figs \ref{fig:analysis} and \ref{fig:contrast_speed}.

{\Large {\sc {\bf (\fig \ref{fig:overview} near here.) }}}

{\Large {\sc {\bf (\fig \ref{fig:flowchart} near here.) }}}

{\Large {\sc {\bf (\fig \ref{fig:analysis} near here.) }}}

\subsection{Related Work}\label{related_work}

Several systems have allowed manual or manually-assisted digitization
of the trajectories of freely flying animals. In the 1970s, Land and
Collett performed pioneering studies on the visual guidance of flight
in blowflies (\citeyear{Land_Collett_1974}) and later, in hoverflies
\citep{Collett_Land_1975, Collett_Land_1978}. By the end of the 70s
and into the 80s, 3D reconstructions using two views of flying insects
were performed \citep{wehrhahn_1979, buelthoff_1980, wehrhahn_1982,
  Dahmen_1984, Wagner_1986}. In one case, the shadow of a bee on a
planar white surface was used as a second view to perform 3D
reconstruction \citep{Srinivasan_2000}. Today, hand digitization is
still used when complex kinematics, such as wing shape and position,
are desired, such as in {\em Drosophila} \citep{Fry_2003}, cockatoo
\citep{Hedrick_2007} and bats \citep{Tian_2006}.

Several authors have solved similar automated multi-target tracking
problems using video. For example, \citet{Khan05mcmc-basedparticle}
tracked multiple, interacting ants in 2D from a single view using
particle filtering with a Markov Chain Monte Carlo sampling step to
solve the multi-target tracking problem. Later work by the same
authors \citep{Khan06dellaert:mcmc} achieved realtime speeds through
the use of sparse updating techniques. \citet{Branson_2009} addressed
the same problem for walking flies. Their technique uses background
subtraction and clustering to detect flies in the image, and casts the
data association problem as an instance of minimum weight bipartite
perfect matching. In implementing flydra, we found the simpler system
described here to be sufficient for tracking position of flying flies
and hummingbirds (see Section \ref{implementation}). In addition to
tracking in three dimensions rather than two, a key difference between
the work described about and those addressed in the present work is
that the interactions between our animals are relatively weak (see
Section \ref{data_association}, especially equation
\ref{eqn:data_association_approx}), and we did not find it necessary
to implement a more advanced tracker. Nevertheless, the present work
could be used as the basis for a more advanced tracker, such as one
using a particle filter \citep[e.g.][]{Klein_2008}. In that case, the
posterior from the Extended Kalman Filter (see Section
\ref{kalman_filtering}) could be used as the proposal distribution for
the particle filter. Others have decentralized the multiple object
tracking problem to improve performance, especially when dealing with
dynamic occlusions due to targets occluding each other
\citep[e.g.][]{Qu_2007a,Qu_2007b}. Additionally, tracking of dense
clouds of starlings
\citep{Ballerini_2008,Cavagna_2008a,Cavagna_2008b,Cavagna_2008c} and
fruit flies \citep{Wu_2009,Zou_2009} has enabled detailed
investigation of swarms, although these systems are currently
incapable of operating in realtime.  By filming inside a corner-cube
reflector, multiple (real and reflected) images allowed
\citet{Bomphrey_2009} to track flies in 3D with only a single camera,
and the tracking algorithm presented here could make use of this
insight.

Completely automated 3D animal tracking systems have more recently
been created such as systems with two cameras that track flies in
realtime \citep{Marden_1997,Fry_2004,Fry_2008}. The system of
\citet{Grover_2008}, similar in many respects to the one we describe
here, tracks the visual hull of flies using three cameras to
reconstruct a polygonal model of the 3D shape of the flies. Our
system, briefly described in a simpler, earlier form in
\citet{Maimon_2008}, differs in several ways. First, flydra has a
design goal of tracking over large volumes, and, as a result of the
associated limited spatial resolution (rather than due to a lack of
interest), flydra is concerned only with the location and orientation
of the animal. Second, to facilitate tracking over large volumes, the
flydra system uses a data association step as part of the tracking
algorithm. The data association step allows flydra to deal with
additional noise (false positive feature detections) when dealing with
low contrast situations often present when attempting to track in
large volumes.  Third, our system utilizes a general multi view
geometry, whereas the system of \citet{Grover_2008} utilizes epipolar
geometry, limiting its use to three cameras.  Finally, although their
system operates in realtime, no measurements of latency were provided
by \citet{Grover_2008} with which to compare our measurements.

\subsection{Notation}

In the equations to follow, letters in a bold, roman font signify a
vector, which may be specified by components enclosed in parentheses
and separated by commas. Matrices are written in roman font with
uppercase letters. Scalars are in italics. Vectors always act like a
single column matrix, such that for vector $\genericvector = (a,b,c)$,
the multiplication with matrix $\genericmatrix$ is $\genericmatrix
\genericvector = \genericmatrix \begin{bmatrix}
  \genericvector \end{bmatrix} = \genericmatrix \begin{bmatrix} a \\ b
  \\ c \end{bmatrix} $.

\section{2D feature extraction}\label{2D_features}

The first stage of processing converts digital images into a list of
feature points using an elaboration of a background subtraction
algorithm. Because the image of a target is usually only a few pixels
in area, an individual feature point from a given camera characterizes
that camera's view of the target. In other words, neglecting missed
detections or false positives, there is usually a one-to-one
correspondence between targets and extracted feature points from a
given camera. Nevertheless, our system is capable of successful
tracking despite missing observations due to occlusion or low contrast
(see Section \ref{kalman_filtering}) and rejecting false positive
feature detections (as described in Section \ref{data_association}).

In the Bayesian framework, all feature points for time $t$ are the
observation $\allZee_t$. The $i$th of $\numcams$ cameras returns
$\numtwodpoints$ feature points, with each point $\zee_{ij}$ being a
vector $\zee_{ij}=(u,v,\alpha,\beta, \theta,\epsilon)$ where $u$ and
$v$ are the coordinates of the point in the image plane and the
remaining components are local image statistics described
below. $\allZee_t$ thus consists of all such feature points for a
given frame $\allZee_t = \{
\zee_{11},\ldots,\zee_{1m},\ldots,\zee_{n1},\ldots,\zee_{nm}\}$. (In
the interest of simplified notation, our indexing scheme is slightly
misleading here --- there may be varying numbers of features for each
camera rather than always $m$ as suggested.)

The process to convert a new image to a series of feature points uses
a process based on background subtraction using the running Gaussian
average method \citep[reviewed in][]{Piccardi_2004}. To achieve fast
image processing required for realtime operation, many of these
operations are performed utilizing the high-performance {\em single
  instruction multiple data} (SIMD) extensions available on recent x86
CPUs. Initially, an absolute difference image is made, where each
pixel is the absolute value of the difference between the incoming
frame and the background image. Feature points that exceed some
threshold difference from the background image are noted and a small
region around each pixel is subjected to further analysis.  For the
$j$th feature, the brightest point has value $\beta_j$ in this
absolute difference image. All pixels below a certain fraction
(e.g. 0.3) of $\beta_j$ are set to zero to reduce moment arms caused
by spurious pixels. Feature area $\alpha_j$ is found from the 0th
moment, the feature center $(\tilde{u}_j,\tilde{v}_j)$ is calculated
from the 1st moment and the feature orientation $\theta_j$ and
eccentricity $\epsilon_j$ are calculated from higher moments. After
correcting for lens distortion (see Section \ref{calibration}), the
feature center is $(u_j,v_j)$. Thus, the $j$th point is characterized
by the vector $\zee_j = (u_j,v_j,\alpha_j,\beta_j, \theta_j,
\epsilon_j)$. Such features are extracted on every frame from every
camera, although the number of points $\numtwodpoints$ found on each
frame may vary. We set the initial thresholds for detection low to
minimize the number of missed detections---false positives at this
stage are rejected later by the data association algorithm (Section
\ref{data_association}).

Our system is capable of dealing with illumination conditions that
vary slowly over time by using an ongoing estimate of the background
luminance and its variance, which are maintained on a per-pixel basis
by updating the current estimates with data from every 500th frame (or
other arbitrary interval). A more sophisticated 2D feature extraction
algorithm could be used, but we have found this scheme to be
sufficient for our purposes and sufficiently simple to operate with
minimal latency.

While the realtime operation of flydra is essential for experiments
modifying sensory feedback, another advantage of an online tracking
system is that the amount of data required to be saved for later
analysis is greatly reduced. By performing only 2D feature extraction
in realtime, to reconstruct 3D trajectories later, only the vectors
$\zee_j$ need be saved, resulting in orders of magnitude less data
than the full-frame camera images. Thus, to achieve solely the low
data rates of realtime tracking, the following sections dealing with
3D are not necessary to be implemented for this benefit of realtime
use. Furthermore, raw images taken from the neighborhood of the
feature points could also be extracted and saved for later analysis,
saving slightly more data, but still at rates substantially less than
the full camera frames provide. This fact is particularly useful for
cameras with a higher data rate than hard drives can save, and such a
feature is implemented in flydra.

{\Large {\sc {\bf (\fig \ref{fig:humdra} near here.) }}}

\fig \ref{fig:humdra} shows the parameters $(u,v,\theta)$ from the 2D
feature extraction algorithm during a hummingbird flight. These 2D
features, in addition to 3D reconstructions, are overlaid on raw
images extracted and saved using the realtime image extraction
technique described above.

\section{Multi target tracking}\label{tracking}

The goal of flydra, as described in Section \ref{algo_overview}, is to
find the MAP estimate of the state of all targets. For simplicity, we
model interaction between targets in a very limited way. Although in
many cases the animals we are interested in tracking do interact (for
example, hummingbirds engage in competition in which they threaten or
even contact each other), mathematically limiting the interaction
facilitates a reduction in computational complexity. First, the
process update is independent for each $k$th animal
\begin{equation}
p(\allstate_t \vert \allstate_{t-1}) = \prod_k p(\allstate_{t,k} \vert \allstate_{t-1,k}).
\end{equation} Second, we implemented only a slight coupling between targets in the
data association algorithm. Thus, the observation likelihood model
$p(\allZee_{t} \vert \allstate_t )$ is independent for each target
with the exception described in Section \ref{track_merging}, and
making this assumption allows use of the Nearest Neighbor Standard
Filter as described below.

Modeling individual target states as mostly independent allows the
problem of estimating the MAP of joint target state $\allstate$ to be
treated nearly as $l$ independent, smaller problems. One benefit of
making the assumption of target independence is that the target
tracking and data association parts of our system are
parallelizable. Although not yet implemented in parallel, our system
is theoretically capable of tracking very many (tens or hundreds)
targets simultaneously with low latency on a computer with
sufficiently many processing units.

The cost of this near-indepdence assumption is reduced tracking
accuracy during periods of near contact (see Secton
\ref{track_merging}). Data from these periods could be analyzed later
using more sophisticated multi-target tracking data association
techniques, presumably in an offline setting, especially because such
periods could be easily identified using a simple algorithm. All data
presented in this paper utilized the procedure described here.

\subsection{Kalman filtering}\label{kalman_filtering}

The standard Extended Kalman Filter (EKF) approximately estimates
statistics of the posterior distribution (equation
\eqnref{eqn:simple_bayes}) for non-linear processes with additive
Gaussian noise (details are given in \ref{EKF}). To utilize this
framework, we make the assumption that noise in the relevant processes
is Gaussian. Additionally, our target independence assumption allows a
single Kalman filter implementation to be used for each tracked
target. The EKF estimates state and its covariance based on a prior
state estimate and incoming observations by using models of the state
update process, the observation process, and estimates of the noise of
each process.

We use a linear model for the dynamics of the system and a nonlinear
model of the observation process.  Specifically, the time evolution of
the system is modeled with the linear discrete stochastic model
\begin{equation}\label{kalman_state_eqn}
\kstatevec_t = \kalmanA \kstatevec_{t-1} + \kalmanw .
\end{equation}
We treat the target as an orientation-free particle, with state vector
$\kstatevec = (x,y,z,\dot{x},\dot{y},\dot{z})$ describing position and
velocity in 3D space. The process update model $\kalmanA$ represents,
in our case, the laws of motion for a constant velocity particle
$$ \kalmanA = \begin{bmatrix} 1 & 0 & 0 & dt & 0 & 0 \\ 0 & 1 & 0 & 0
  & dt & 0 \\ 0 & 0 & 1 & 0 & 0 & dt \\ 0 & 0 & 0 & 1 & 0 & 0 \\ 0 & 0
  & 0 & 0 & 1 & 0 \\ 0 & 0 & 0 & 0 & 0 & 1\end{bmatrix} $$ with $dt$
being the time step. Maneuvering of the target (deviation from the
constant velocity) is modeled as noise in this formulation. The random
variable $\kalmanw$ represents this process update noise with a normal
probability distribution with zero mean and the process covariance
matrix $\kalmanQ$. Despite the use of a constant velocity model, the
more complex trajectory of a fly or other target is accurately
estimated by updating the state estimate with frequent observations.

\begin{figure*}
\includegraphics[width=\onecolfig]{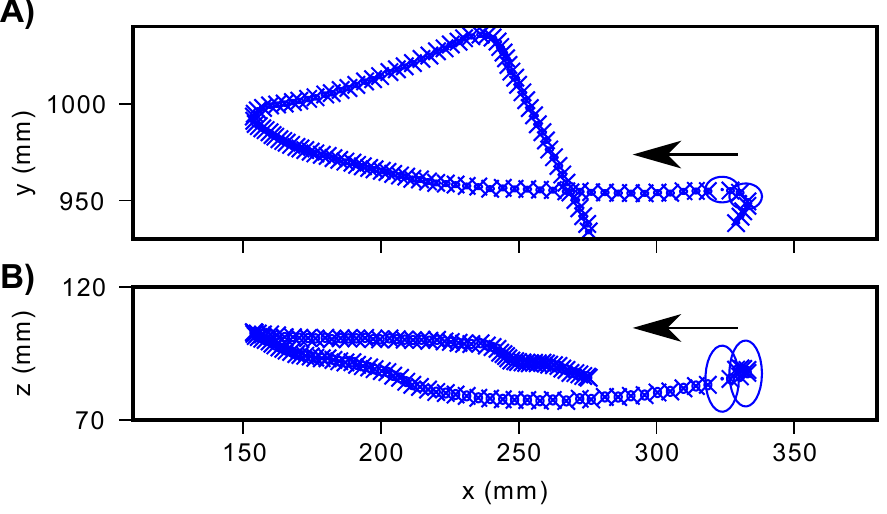}

\caption{Two seconds of a reconstructed {\em Drosophila} trajectory.
  A) Top view. B) Side view of same trajectory.  Kalman filter based
  estimates of fly position $\kstatevec$ are plotted as dots at the
  center of ellipsoids, which are the projections of the multivariate
  normal specified by the covariance matrix $\kesterr$. Additionally,
  position estimated directly by triangulation of 2D point locations
  (see \ref{triangulation}) is plotted with crosses. Fly began on the
  right and flew in the direction denoted by the arrow. Note that for
  two frames near the beginning, only a single camera contributed to
  the tracking and the error estimate increased. }

\label{fig:ellipsoids}

\end{figure*}

{\Large {\sc {\bf (\fig \ref{fig:ellipsoids} near here.) }}}

For a given data association hypothesis (see Section
\ref{data_association}), a set of observations is available for target
$k$. A nonlinear observation model, requiring use of an extended
Kalman filter, is used to describe the action of a projective camera
(equations
\ref{kalman_observation_equation}--\eqnref{eqn:observation_func}
below). This allows observation error to be modeled as Gaussian noise
on the image plane. Furthermore, during tracking, triangulation
happens only implicitly, and error estimates of target position are
larger along the direction of the ray between the target and the
camera center. (To be clear, explicit triangulation {\em is} performed
during the initialization of a Kalman model target, as explained in
Section \ref{data_association}.) Thus, observations from alternating
single cameras on successive frames would be sufficient to produce a
3D estimate of target position. For example, \fig \ref{fig:ellipsoids}
shows a reconstructed fly trajectory in which two frames were lacking
data from all but one camera. During these two frames, the estimated
error increased, particularly along the camera-fly axis, and no
triangulation was possible. Nevertheless, an estimate of 3D position
was made and appears reasonable. The observation vector $\kobsvec =
(u_1,v_1,u_2,v_2,...,u_n,v_n)$ is the vector of the
distortion-corrected image points from $\numcams$ cameras. The
non-linear observation model relates $\kobsvec_t$ to the state
$\kstatevec_t$ by
\begin{equation}\label{kalman_observation_equation}
\kobsvec_t = \kalmanh(\kstatevec_t) + \kalmanv
\end{equation}
where $\kstatevec_t$ is the state at time $t$, the function $\kalmanh$
models the action of the cameras and $\kalmanv$ is observation noise.
The vector-valued $\kalmanh(\kstatevec)$ is the concatenation of the
image points found by applying the image point equations (equations
\eqnref{eqn:dehomogenize} and \eqnref{imagepoint}) to each of the
$\numcams$ cameras
\begin{equation}
\begin{split}
\kalmanh(\kstatevec)=(\kalmanh_1(\kstatevec), ...,
\kalmanh_n(\kstatevec)) \label{eqn:observation_func} \\
= (\Eu_1,\Ev_1,...,\Eu_n,\Ev_n) \\
= (\Er_1/\Et_1,\Es_1/\Et_1,...,\Er_n/\Et_n,\Es_n/\Et_n) \\
= (\dehomogenize(\cammatrix_1\threedeevec), ..., \dehomogenize(\cammatrix_n\threedeevec)).
\end{split}
\end{equation}
The overbar ($\bar{}$) denotes a noise-free prediction to which the
zero-mean noise vector $\kalmanv$ is added, and $\threedeevec$ is the
homogeneous form of the first three components of $\kstatevec$. The
random variable $\kalmanv$ models the observation noise as normal in
the image plane with zero mean and covariance matrix
$\kalmanR$.

At each time step $t$, the extended Kalman filter formulation is then
used to estimate the state $\kstateest$ in addition to the error
$\kesterr$ (see \ref{EKF}). Together, the data associated with each
target is $\Gamma = \{ \kstateest, \kesterr \}$. With the possibility of
multiple targets being tracked simultaneously, the $k$th target is
assigned $\Gamma^k$.

One issue we faced when implementing the Kalman filter was parameter
selection. Our choice of parameters was done through educated guesses
followed by an iterated trial-and-error procedure using several
different trajectories' observations. The parameters that resulted in
trajectories closest to those seen by eye and with least magnitude
error estimate $\kesterr$ were used.  We obtained good results, for
fruit flies measured under the conditions of our setups, with the
process covariance matrix $\kalmanQ$ being diagonal, with the first
three entries being 100 mm$^2$ and the next three being 0.25
(m/sec)$^2$. Therefore, our model treats maneuvering as position and
velocity noise. For the observation covariance matrix $\kalmanR$, we
found good results with a diagonal matrix with entries of 1,
corresponding to variance of the observed image positions of one
pixel. Parameter selection could be automated by an
expectation-maximization algorithm, but we found this was not
necessary.

Another issue is missing data---in some time steps, all views of the
fly may be occluded or low contrast, leaving a missing value of
$\kobsvec$ for that time step. In those cases, we simply set the {\em
  a posteriori} estimate to the {\em a priori} prediction, as follows
from equation \eqnref{eqn:simple_bayes}. In these circumstances, the
error estimate $\kesterr$ grows by the process covariance $\kalmanQ$,
and does not get reduced by (nonexistant) new observations. This
follows directly from the Kalman filter equations (\ref{EKF}). If too
many successive frames with no observations occur, the error estimate
will exceed a threshold and tracking will be terminated for that
target (described in Section \ref{birth_model}).

\subsection{Data association}\label{data_association}

One simplification made in the system overview (Section
\ref{algo_overview}) was to neglect the data association problem---the
assignment of observations to targets. We address the problem by
marginalizing the observation likelihood across hidden data
association variables $\dassocvar$, where each $\dassocvar$
corresponds to a different hypothesis about how the feature points
correspond with the targets. Thus, the model of observation likelihood
from equation \eqnref{eqn:simple_bayes} becomes
\begin{equation} \label{eqn:marginalize_data_association}
p(\allZee_{t} \vert \allstate_t ) = \sum_\dassocvar p(\allZee_{t},\dassocvar \vert \allstate_t ) .
\end{equation}

In fact, computing probabilities across all possible data association
hypotheses $\dassocvar$ across multiple time steps would result in a
combinatorial explosion of possibilities. Among the various means of
limiting the amount of computation required by limiting the number of
hypotheses considered, we have chosen a simple method, the Nearest
Neighbor Standard Filter (NNSF) data association algorithm run on each
target independently \citep{bar_shalom_fortmann_1988}. This algorithm
is sufficiently efficient to operate in realtime for typical
conditions of our system. Thus, we approximate the sum of all data
association hypotheses with the single best hypothesis
$\dassocvar_\mathrm{NNSF}$, defined to be the NNSF output for each of
the $k$ independent targets
\begin{equation} \label{eqn:data_association_approx}
p(\allZee_{t},\dassocvar_\mathrm{NNSF} \vert \allstate_t ) \approx
\sum_\dassocvar p(\allZee_{t},\dassocvar \vert \allstate_t ) .
\end{equation}
This implies that we assume hypotheses other than
$\dassocvar_\mathrm{NNSF}$ have vanishingly small probability. Errors
due to this assumption being false could be corrected in a later,
offline pass through the data keeping track of more data association
hypotheses using other algorithms.

$\dassocvar$ is a matrix with each column being the data association
vector for target $k$ such that $\dassocvar = [\assignvec^1 \quad
  \ldots \quad \assignvec^k \quad \ldots \quad ]$. This matrix has
$\numcams$ rows (the number of cameras) and $l$ columns (the number of
active targets). The data association vector $\assignvec^k$ for target
$k$ has $\numcams$ elements of value null or index $j$ of the feature
$\zee_j$ assigned to that target. As described below (Section
\ref{using_incoming_data}), these values are computed from the
predicted location of the target and the features returned from the
cameras.

\subsubsection{Preventing track merging}\label{track_merging}

One well known potential problem with multi target tracking is the
undesired merging of target trajectories if targets begin to share the
same observations.  Before implementing the following rule, flydra
would sometimes suffer from this merging problem when tracking
hummingbirds engaged in territorial competition. In such fights, male
hummingbirds often fly directly at each other and come in physical
contact. To prevent the two trajectories from merging in such cases, a
single pass is made through the data association assignments after
each frame. In the case that more than one target was assigned the
exact same subset of feature points, a comparison is made between the
observation and the predicted observation. In this case, only the
target corresponding to the closest prediction is assigned the data,
and the other target is updated without any observation. We found this
procedure to require minimal additional computational cost, while
still being effective in preventing trajectory merging.

\subsubsection{NNSF and generative model of image features}\label{generative_model}

To implement the NNSF algorithm, we implement a generative model of
feature appearance based on the prior estimate of target state. By
predicting target position in an incoming frame based on prior
information, the system selects 2D image points as being likely to
come from the target by gating unlikely observations, thus limiting
the amount of computation performed.

\begin{figure*}
\includegraphics[width=\onefivecolfig]{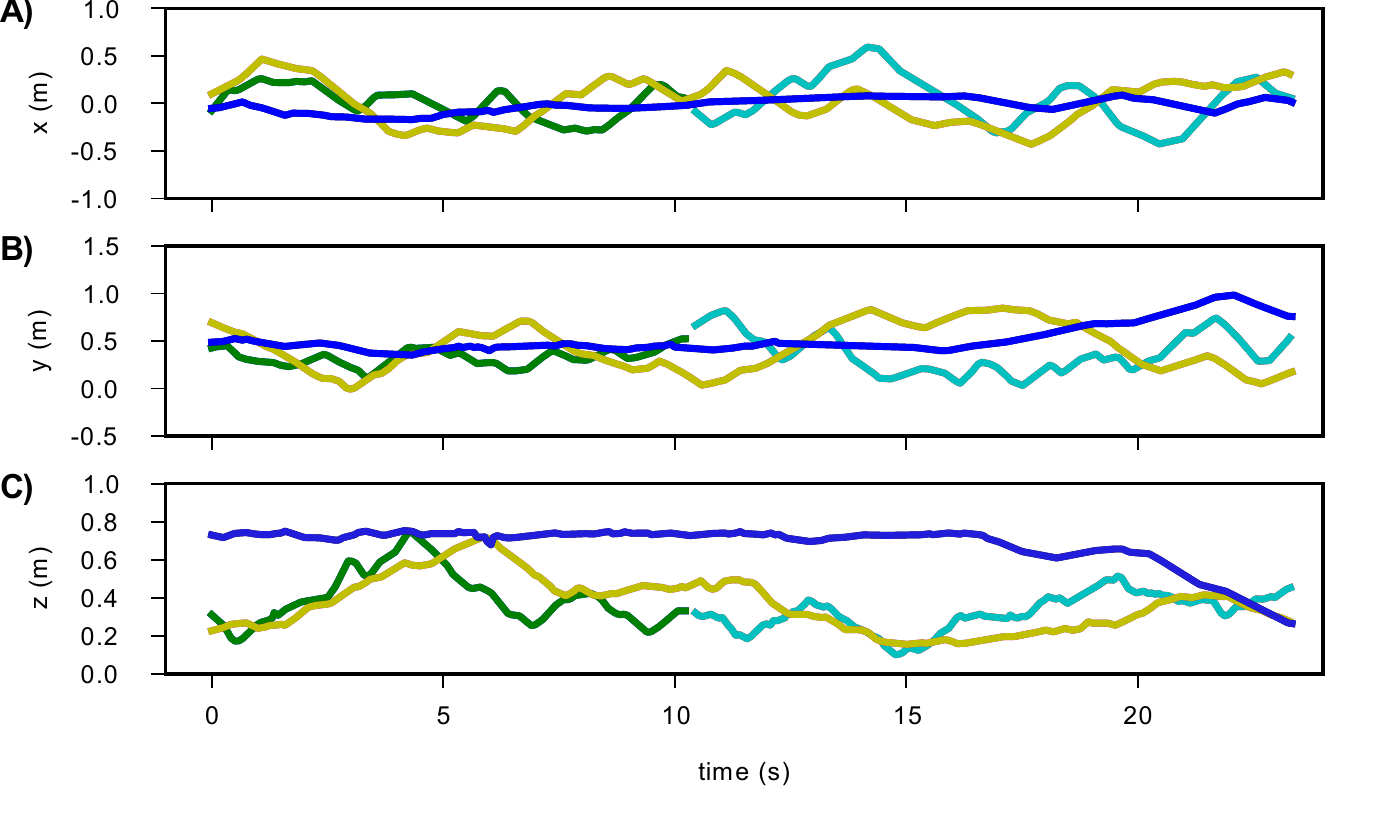}

\caption{Multiple flies tracked simultaneously. Each automatically
  segmented trajectory is plotted in its own color. Note that the dark
  green and cyan trajectories probably came from the same fly which,
  for a period near the 10th second, was not tracked due to a series
  of missed detections or leaving the tracking volume (see Section
  \ref{data_association}).  A) First horizontal axis (x). B) Second
  horizontal axis (y). C) Vertical axis (z). }

\label{fig:multifly}

\end{figure*}

{\Large {\sc {\bf (\fig \ref{fig:multifly} near here.) }}}

Recall from Section \ref{2D_features} that for each time $t$ and
camera $i$, $\numtwodpoints$ feature points are found with the $j$th
point being $\zee_j = (u_j,v_j,\alpha_j,\beta_j, \theta_j,
\epsilon_j)$. The distortion-corrected image coordinates are $(u,v)$,
while $\alpha$ is the area of the object on the image plane measured
by thresholding of the difference image between the current and
background image, $\beta$ is an estimate of the maximum difference
within the difference image, and $\theta$ and $\epsilon$ are the slope
and eccentricity of the image feature.  Each camera may return
multiple candidate points per time step, with all points from the
$i$th camera represented as $\Zee_i$, a matrix whose columns are the
individual vectors $\zee_j$, such that $\Zee_i = \begin{bmatrix}
  \zee_1 & ... & \zee_m \end{bmatrix}$.  The purpose of the data
association algorithm is to assign each incoming point $\zee$ to an
existing Kalman model $\Gamma$, to initialize a new Kalman model, or
attribute it to a false positive (a null target). Furthermore, old
Kalman models for which no recent observations exist due to the target
leaving the tracking volume must be deleted. The use of such a data
association algorithm allows flydra to track multiple targets
simultaneously, as in \fig \ref{fig:multifly}, and, by reducing
computational costs, allows the 2D feature extraction algorithm to
return many points to minimize the number of missed detections.

\subsubsection{Entry and exit of targets}\label{birth_model}

How does our system deal with existing targets losing visibility due
to leaving the tracking volume, occlusion or lowered visual contrast?
What happens when new targets become visible? We treat such occurrences
as part of the update model in the Bayesian framework of Section
\ref{algo_overview}. Thus, in the terminology from that section, our
motion model for all targets $p(\allstate_t \vert \allstate_{t-1})$
includes the possibility of initializing a new target or removing an
existing target. This section describes the procedure followed.

For all data points $\zee$ that remained `unclaimed' by the predicted
locations of pre-existing targets (see Section
\ref{using_incoming_data} below), we use an unguided hypothesis
testing algorithm. This triangulates a hypothesized 3D point for every
possible combination of 2,3,...,$\numcams$ cameras, for a total of
${\numcams \choose 2}$ + ${\numcams \choose 3}$ + ... + ${\numcams
  \choose \numcams}$ combinations.  Any 3D point with reprojection
error less than an arbitrary threshold using the greatest number of
cameras is then used to initialize a new Kalman filter instance.  The
initial state estimate is set to that 3D position with zero velocity
and a relatively high error estimate. Tracking is stopped (a target is
removed) once the estimated error $\kesterr$ exceeds a threshold. This
most commonly happens, for example, when the target leaves the
tracking area and thus receives no observations for a given number of
frames.

\subsubsection{Using incoming data}\label{using_incoming_data}

Ultimately, the purpose of the data association step is to determine
which image feature points will be used by which target. Given the
target independence assumption, each target uses incoming data
independently. This section outlines how the data association
algorithm is used to determine the feature points treated as the
observation for a given target.

To utilize the Kalman filter described in Section
\ref{kalman_filtering}, observation vectors must be formed from the
incoming data.  False positives must be separated from correct
detections, and, because multiple targets may be tracked
simultaneously, correct detections must be associated with a
particular Kalman model. At the beginning of processing for each time
step $t$ for the $k$th Kalman model, a prior estimate of target
position and error $\Gamma^k_{t|t-1}= \{
  \kstateapriori^k,\kesterrapriori^k \}$ is available. It must be
determined which, if any, of the $\numtwodpoints$ image points from
the $i$th camera is associated with the $k$th target.  Due to the
realtime requirements for our system, flydra gates incoming detections
on simple criteria before performing more computationally intensive
tasks.

For target $k$ at time $t$, the data association function $g$ is
\begin{equation}
\assignvec_t^k=g( \Zee_1, ..., \Zee_\numcams,\, \Gamma_{t|t-1}^k).
\end{equation} This is a function of the image points $\Zee_i$
from each of the $\numcams$ cameras and the prior information for
target $k$. The assignment vector for the $k$th target,
$\assignvec^k$, defines which points from which cameras view a
target. This vector has a component for each of the $\numcams$
cameras, which is either null (if that camera doesn't contribute) or
is the column index of $\Zee_i$ corresponding to the associated point.
Thus, $\assignvec^k$ has length $\numcams$, the number of cameras, and
no camera may view the same target more than once. Note, the $k$ and
$t$ superscript and subscript on $\assignvec$ indicate the assignment
vector is for target $k$ at time step $t$, whereas below (Equation
\ref{eqn:assign_vec_component}), the subscript $i$ is used to indicate
the $i$th component of the vector  $\assignvec$.

The data association function $g$ may be written in terms of the
components of $\assignvec$. The $i$th component is the index of the
columns of $\Zee_i$ that maximizes likelihood of the observation given
the predicted target state and error and is defined to be
\begin{equation} \label{eqn:assign_vec_component}
\assignvec_i = \argmax_{j}
( p(\zee_j | \Gamma )), \quad \zee_j \in \Zee_i .
\end{equation} Our likelihood function gates detections
based on two conditions. First, the incoming detected location
$(u_j,v_j)$ must be within a threshold Euclidean distance from the
estimated target location projected on the image. The Euclidean
distance on the image plane is
\begin{equation}
{\rm{dist2d}} = d_{euclid}( \begin{bmatrix} u_j \\ v_j \end{bmatrix},
\dehomogenize(\cammatrix_i \threedeevec) )
\end{equation} where $\dehomogenize(\cammatrix_i \threedeevec)$ finds the
projected image coordinates of $\threedeevec$, where $\threedeevec$ is
the homogeneous form of the first three components of $\kstatevec$,
the expected 3D position of the target. The function $\dehomogenize$
and camera matrix $\cammatrix_i$ are described in \ref{triangulation}.
The gating can be expressed as an indicator function
\begin{equation}\label{eqn:dist_2d_thresh}
\indicator_{\rm{dist2d}}(u_j,v_j) = \left \{ \begin{matrix}
1 &\mbox{if}\ \rm{dist2d} < {\rm thresh}_{\rm{dist2d}}, \\
0 &\mbox{otherwise}.
\end{matrix}\right.
\end{equation}
Second, the area of the detected object ($\alpha_j$) must be greater
than a threshold value, expressed as \begin{equation} \label{eqn:area_thresh}
\indicator_{\rm{area}}(\alpha_j) = \left \{ \begin{matrix}
1 &\mbox{if}\ \alpha_j > {\rm thresh}_{\rm{area}}, \\
0 &\mbox{otherwise}.
\end{matrix}\right.
\end{equation} If these conditions are met, the distance of
the ray connecting the camera center and 2D point on the image plane
$(u_j,v_j)$ from the expected 3D location $\vecabar$ is used to
further determine likelihood. We use the Mahalanobis distance, which
for a vector $\veca$ with an expected value of $\vecabar$ with
covariance matrix $\Sigma$ is
\begin{equation} \label{eqn:dist_mahal}
d_{mahal}(\veca,\vecabar) = \sqrt{ (\veca - \vecabar)^\trnsps
  \Sigma^{-1} (\veca - \vecabar) }. \end{equation} Because the
distance function is convex for a given $\vecabar$ and $\Sigma$, we
can solve directly for the closest point on the ray, by setting
$\vecabar$ equal to the first three terms of $\kstateapriori$ and
$\Sigma$ to the upper left $3\cross3$ submatrix of
$\kesterrapriori$. Then, if the ray is a parametrized line of the form
$\linefunc(s)=s\cdot(a,b,c) + (x,y,z)$ where $(a,b,c)$ is the
direction of the ray formed by the image point $(u,v)$ and the camera
center and $(x,y,z)$ is a point on the ray, we can find the value of
$s$ for which the distance between $\linefunc(s)$ and $\vecabar$ is
minimized by finding the value of $s$ where derivative of
$d_{mahal}(\linefunc(s),\;\vecabar)$ is zero. If we call this closest
point $\veca$ and combine equations
\eqnref{eqn:dist_2d_thresh}--\eqnref{eqn:dist_mahal}, then our
likelihood function is
\begin{equation}
 p(\zee_j | \Gamma ) = \indicator_{\rm{dist2d}}(u_j,v_j) \quad  \indicator_{\rm{area}}(\alpha_j) \quad e^{ -  d_{mahal}( \veca, \vecabar ) } \; .
\end{equation}
Note that, due to the multiplication, if either of the first two
factors are zero, the third (and more computationally expensive)
condition need not be evaluated.

\section{Camera and lens calibration}\label{calibration}

Camera calibrations may be obtained in any way that produces camera
calibration matrices (described in \ref{triangulation}) and,
optionally, parameters for a model of the non-linear distortions of
cameras. Good calibrations are critical for flydra because, as target
visibility changes from one subset of cameras to another subset, any
misalignment of the calibrations will introduce artifactual movement
in the reconstructed trajectories. Typically, we obtain camera
calibrations in a two step process. First, the Direct Linear
Transformation (DLT) algorithm \citep{abdelaziz_1971} directly
estimates camera calibration matrices $\cammatrix_i$ that could be
used for triangulation as described in \ref{triangulation}. However,
because we use only about 10 manually digitized corresponding 2D/3D
pairs per camera, this calibration is relatively low-precision and, as
performed, ignores optical distortions causing deviations from the
linear simple pinhole camera model. Therefore, an automated
\emph{Multi-Camera Self Calibration Toolbox} \citep{svoboda_2005} is
used as a second step. This toolbox utilizes inherent numerical
redundancy when multiple cameras are viewing a common set of 3D points
through use of a factorization algorithm \citep{Sturm96afactorization}
followed by bundle adjustment (reviewed in Section 18.1 of
\citealp{hartley_zisserman_2003}). By moving a small LED point light
source through the tracking volume (or, indeed, a freely flying fly),
hundreds of corresponding 2D/3D points are generated which lead to a
single, overdetermined solution which, without knowing the 3D
positions of the LED, is accurate up to a scale, rotation, and
translation. The camera centers, either from the DLT algorithm or
measured directly, are then used to find the best scale, rotation, and
translation. As part of the Multi-Camera Self Calibration Toolbox,
this process may be iterated with an additional step to estimate
non-linear camera parameters such as radial distortion
\citep{svoboda_2005} using the \emph{Camera Calibration Toolbox} of
Bouguet. Alternatively, we have also implemented the method of
\citet{Prescott_1997} to estimate radial distortion parameters before
use of Svoboda's toolbox, which we found necessary when using wide
angle lenses with significant radial distortion (e.g. 150 pixels in
some cases).

\section{Implementation and evaluation}\label{implementation}

We built three different flydra systems: a five camera, six computer
100 fps system for tracking fruit flies in a 0.3m x 0.3m x 1.5m arena
(e.g. \figs \ref{fig:overview}, \ref{fig:contrast_speed} and
\citealt{Maimon_2008}, which used the same hardware but a simpler
version of the tracking software), an eleven camera, nine computer 60
fps system for tracking fruit flies in a large---2m diameter x 0.8m
high---cylinder (e.g. \fig \ref{fig:ellipsoids}) and a four camera,
five computer 200 fps system for tracking hummingbirds in a 1.5m x
1.5m x 3m arena (e.g. \fig \ref{fig:humdra}). Apart from the low-level
camera drivers, the same software is running on each of these systems.

We used the Python computer language to implement flydra. Several
other pieces of software, most of which are open source, are
instrumental to this system: motmot \citep{straw_2009b}, pinpoint,
pytables, numpy, scipy, Pyro, wxPython, tvtk, VTK, matplotlib,
PyOpenGL, pyglet, Pyrex, cython, ctypes, Intel IPP, ATLAS, libdc1394,
PTPd, gcc, and Ubuntu. We used Intel Pentium 4 and Core 2 Duo based
computers.

The accuracy of 3D reconstructions was verified in two ways. First,
the distance between 2D points projected from a 3D estimate derived
from the originally extracted 2D points is a measure of calibration
accuracy. For all figures shown, the mean reprojection error was less
than one pixel, and for most cameras in most figures, was less than
0.5 pixels. Secondly, the 3D coordinates and distances between
coordinates measured through triangulation were verified against
values measured physically. For two such measurements in the system
shown in \fig \ref{fig:overview}, these values were within 4 percent.

\begin{figure*}
\includegraphics[width=\onecolfig]{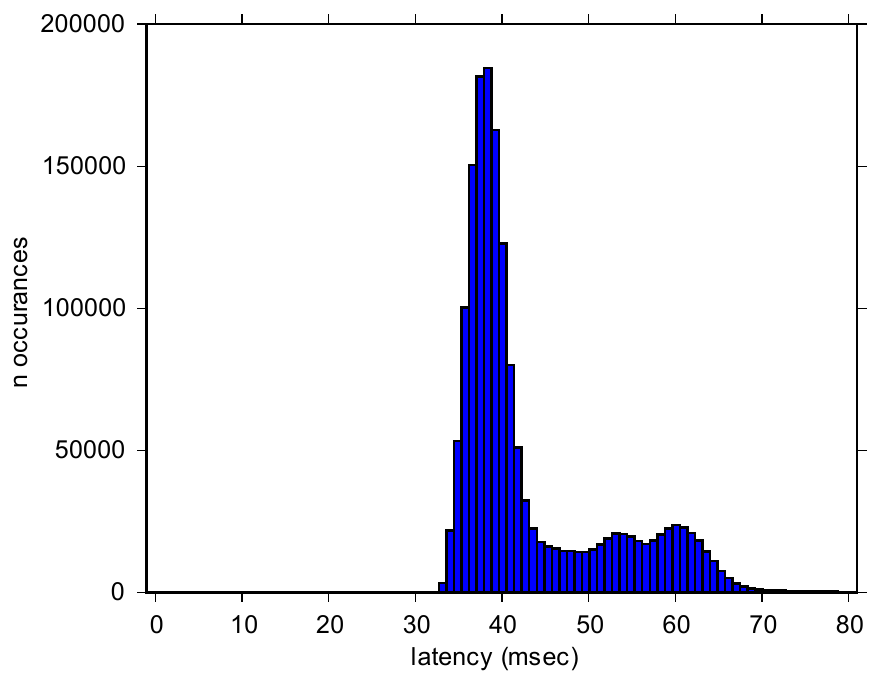}
\caption{Latency of one tracking system. A histogram of the latency of
  3D reconstruction was generated after tracking 20 flies for 18
  hours. Median latency was 39 msec.}
\label{fig:latency}
\end{figure*}

{\Large {\sc {\bf (\fig \ref{fig:latency} near here.) }}}

We measured the latency of the 3D reconstruction by synchronizing
several clocks involved in our experimental setup and then measuring
the duration between onset of image acquisition and completion of the
computation of target position. When flydra is running, the clocks of
the various computers are synchronized to within 1 microsecond by
PTPd, the precise time protocol daemon, an implementation of the IEEE
1588 clock synchronization protocol
\citep{Correll_2005}. Additionally, a microcontroller (AT90USBKEY,
Atmel, USA) running custom firmware is connected over USB to the
central reconstruction computer and is synchronized using an algorithm
similar to PTPd, allowing the precise time of frame trigger events to
be known by processes running within the computers. Measurements were
made of the latency between the time of the hardware trigger pulse
generated on the microcontroller to start acquisition of frame $t$ and
the moment the state vector $\kstateapost$ was computed. These
measurements were made with a central computer being a 3 GHz Intel
Core 2 Duo CPU. As shown in \fig \ref{fig:latency}, the median 3D
reconstruction timestamp is $39$ msec. Further investigation showed
the makeup of this delay.  From the specifications of the cameras and
bus used, $19.5$ msec is a lower bound on the latency of transferring
the image across the IEEE 1394 bus (and could presumably be reduced by
using alternative technologies such as Gigabit Ethernet or Camera
Link). Further measurements on this system showed that 2D feature
extraction takes 6--12 msec, and that image acquisition and 2D feature
extraction together take 26--32 msec. The remainder of the latency to
3D estimation is due to network transmission, triangulation, and
tracking, and, most likely, non-optimal queueing and ordering of data
while passing it between these stages. Further investigation and
optimization has not been performed.

\section{Experimental Possibilities}

A few examples serve to illustrate some of the capabilities of
flydra. We are actively engaged in understanding the sensory-motor
control of flight in the fruit fly {\em Drosophila melanogaster}. Many
basic features of the effects of visual stimulation on the flight of
flies are known, and the present system allows us to characterize
these phenomena in substantial detail. For example, the presence of a
vertical landmark such as a black post on a white background greatly
influences the structure of flight, causing flies to `fixate', or turn
towards, the post \citep{Kennedy_1939}. By studying such behavior in
free flight (e.g. \fig \ref{fig:analysis}), we have found that flies
approach the post until some small distance is reached, and then often
turn rapidly and fly away.

\begin{figure*}
\includegraphics[width=\onefivecolfig]{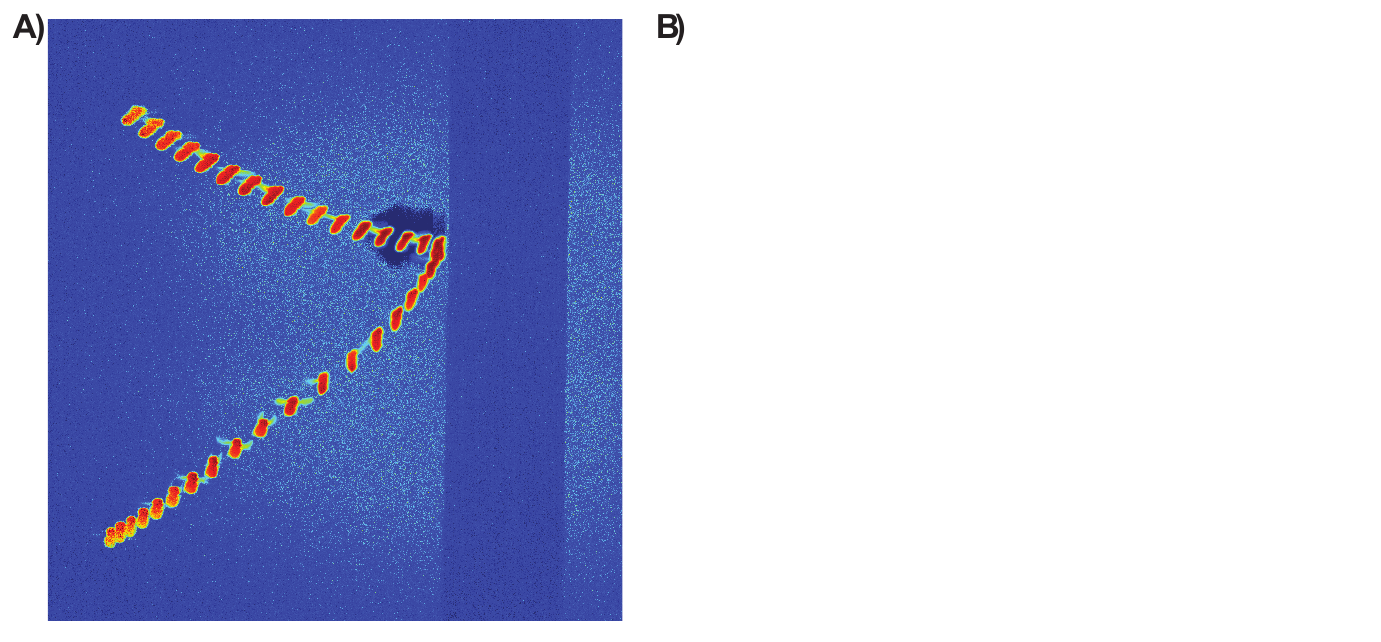}
\caption{Example applications for flydra. A) By tracking flies in
  real-time, we can trigger high speed video cameras (6000 fps) to
  record interesting events for later analysis.  B) A reconstruction
  of the visual scene viewed through the left eye of
  \emph{Drosophila}.  Such visual reconstructions can be used to
  simulate neural activity in the visual system
  \citep[e.g.][]{neumann_2002,dickson_2008}.}
\label{fig:applications}
\end{figure*}

{\Large {\sc {\bf (\fig \ref{fig:applications} near here.) }}}

Because we have an online estimate of fly velocity and position in
$\kstateest$, we can predict the future location of the fly. Of
course, the quality of the prediction declines with the duration of
extrapolation, but it is sufficient for many tasks, even with the
latency taken by the 3D reconstruction itself. One example is the
triggering of high resolution, high speed cameras (e.g. 1024x1024
pixels, 6000 frames per second as shown in \fig
\ref{fig:applications}A). Such cameras typically buffer their images
to RAM and are downloaded offline. We can construct a trigger
condition based on the position of an animal (or a recent history of
position, allowing triggering only on specific maneuvers). \fig
\ref{fig:applications}A shows a contrast-enhancing false color montage
of a fly makes a close approach to a post before leaving. By studying
precise movements of the wings and body in addition to larger scale
movement through the environment, we are working to understand the
neural and biomechanical mechanisms underlying control of flight.

An additional possibility enabled by low latency 3D state estimates is
that visual stimuli can be modulated to produce a `virtual reality'
environment in which the properties of the visual feedback loop can be
artificially manipulated
\citep{Fry_2004,Fry_2008,Strauss_1997,straw_2008b}. In these types of
experiments, it is critical that moving visual stimuli do not affect
the tracking. For this reason, we illuminate flies with near-IR light
and use high pass filters in front of the camera lenses (e.g. R-72,
Hoya Corporation). Visual cues provided to the flies are in the
blue-green range.

By estimating the orientation of the animal, approximate
reconstructions of the visual stimulus experienced by the animal may
be made. For example, to make \fig \ref{fig:applications}B, a
simulated view through the compound eye of a fly, we assumed that the
fly head, and thus eyes, was oriented tangent to the direction of
travel, and that the roll angle was fixed at zero. This information,
together with a 3D model of the environment, was used to generate the
reconstruction \citep{neumann_2002,dickson_2008}. Such reconstructions
are informative for understanding the nature of the challenge of
navigating visually with limited spatial resolution. Although it is
known that flies do move their head relative to their longitudinal
body axis, these movements are generally small \citep{Schilstra_1999},
and thus the errors resulting from the assumptions listed above could
be reduced by estimating body orientation using the method described
in \ref{triangulation}. Because a fly's eyes are fixed to its head,
further reconstruction accuracy could be gained by fixing the head
relative to the body (by gluing the head to the thorax), although the
behavioral effects of such a manipulation would need to be
investigated.

Finally, because of the configurability of the system, it is feasible
to consider large scale tracking in naturalistic environments that
would lead to greater understanding of the natural history of flies
\citep{Stamps_2005} or other animals.

\section{Effect of contrast on speed regulation in {\em Drosophila}}

At low levels of luminance contrast, when differences in the luminance
of different parts of a scene are minimal, it is difficult or
impossible to detect motion of visual features. In flies, which judge
self-motion (in part) using visual motion detection, the exact nature
of contrast sensitivity has been used as a tool to investigate the
fundamental mechanism of motion detection using electrophysiological
recordings. At low contrast levels, these studies have found a
quadratic relationship between membrane potential and luminance
contrast \citep{Dvorak_Srinivasan_French_1980, Egelhaaf_Borst_1989,
  Srinivasan_Dvorak_1980}. This result is consistent with
Hassenstein-Reichardt correlator model for elementary motion detection
(the HR-EMD, \citealp{Hassenstein_Reichardt_1956}), and these findings
are part of the evidence to support the hypothesis that the fly visual
system implements something very similar to this mathematical model.

Despite these and other electrophysiological findings suggesting the
HR-EMD may underlie fly motion sensitivity, studies of flight speed
regulation and other visual behaviors in freely flying flies
\citep{David_1982} and honey bees
\citep{Srinivasan_Lehrer_Kirchner_Zhang_1991,
  Si_Srinivasan_Zhang_2003, Srinivasan_Zhang_Chandrashekara_1993} show that the free-flight
behavior of these insects is inconsistent with a flight velocity
regulator based on a simple HR-EMD model. More recently,
\citet{Baird_Srinivasan_Zhang_Cowling_2005} have shown that over a
large range of contrasts, flight velocity in honey bees is nearly
unaffected by contrast. As noted by those authors, however, their
setup was unable to achieve true zero contrast due to imperfections
with their apparatus. They suggest that contrast adaptation
\citep{Harris_et_al_2000} may have been responsible for boosting the
responses to low contrasts and attenuating responses to high
contrast. This possibility was supported by the finding that forward
velocity was better regulated at the nominal ``zero contrast''
condition than in the presence of an axial stripe, which may have had
the effect of preventing contrast adaptation while provide no contrast
perpendicular to the direction of flight
\citep{Baird_Srinivasan_Zhang_Cowling_2005}.

\begin{figure*}
\includegraphics[width=\onecolfig]{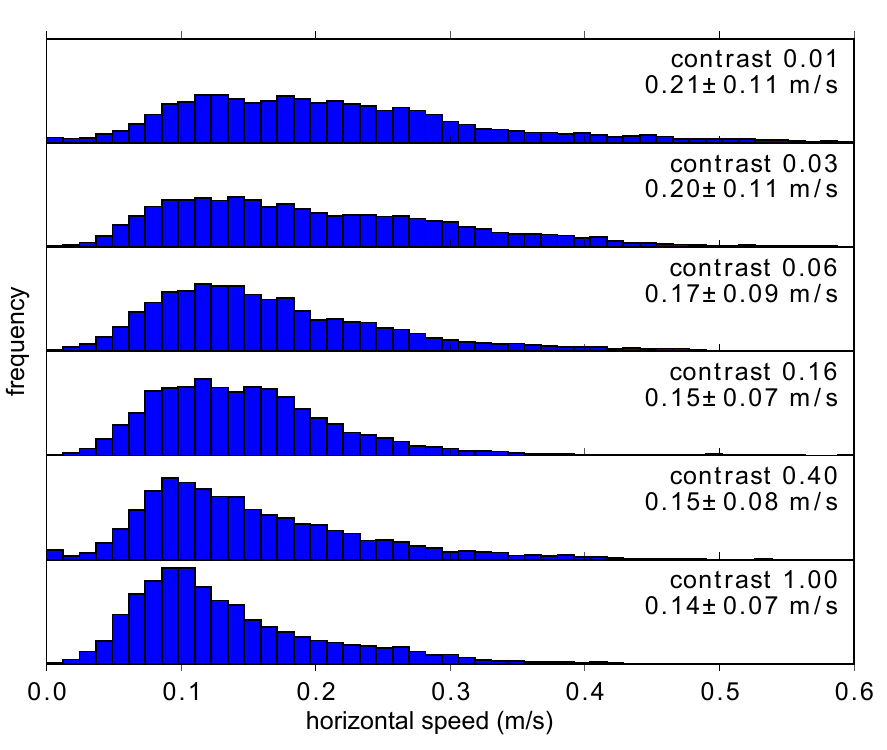}
\caption{ {\em Drosophila melanogaster} maintain a lower flight speed
  with lower variability as visual contrast is increased. Mean and
  standard deviation of flight speeds are shown in text, and each
  histogram is normalized to have equal area.  Ambient illumination
  reflected from one surface after being scattered from other
  illuminated surfaces slightly reduced contrast from the nominal
  values shown here.}
\label{fig:contrast_speed}
\end{figure*}

To test the effect of contrast on the regulation of flight speed in
{\em Drosophila melanogaster}, we performed an experiment, the results
of which are shown in \fig \ref{fig:contrast_speed}. In this setup, we
found that when contrast was sufficiently high (Michelson contrast
$\ge 1.6$), flies regulated their speed to a mean speed of 0.15 m/s
with a standard deviation of 0.07. As contrast was lowered, the mean
speed increased, as did variability of speed, suggesting that speed
regulation suffered due to a loss of visual feedback. To perform these
experiments, a computer projector (DepthQ, modified to remove color
filter wheel, Lightspeed Design, USA) illuminated the long walls and
floor of a 0.3m x 0.3m x 1.5m arena with a regular checkerboard
pattern (5cm squares) of varying contrast and fixed luminance (2
cd/m$^2$). The test contrasts were cycled, with each contrast
displayed for 5 minutes. 20 females flies were released into the arena
and tracked over 12 hours. Any flight segments more than 5 cm from the
walls, floor or ceiling were analyzed, although for the majority of
the time flies were standing or walking on the floors or walls of the
arena. Horizontal flight speed was measured as the first derivative of
position in the XY direction, and histograms were computed with each
frame constituting a single sample. Because the identity of the flies
could not be tracked for the duration of the experiment, the data
contain pseudo-replication --- some flies probably contributed more to
the overall histogram than others.  Nevertheless, the results from 3
separate experimental days with 20 new flies tested on each day were
each qualitatively similar to the pooled results shown, which include
1760 total seconds of flight in which tracked fly was 5 cm or greater
from the nearest arena surface and were acquired during 30 cumulative
hours.

The primary difference between our findings on the effect of contrast
on flight speed in {\em Drosophila melanogaster} compared to that
found in honey bees by \citet{Baird_Srinivasan_Zhang_Cowling_2005} is
that at low contrasts (below 0.16 Michelson contrast), flight speed in
{\em Drosophila} is faster and more variable. This difference could be
due to several non-mutually exclusive factors: 1) Our arena may have
fewer imperfections which create visual contrast. 2) Fruit flies may
have a lower absolute contrast sensitivity than honey bees. 3) Fruit
flies may have a lower contrast sensitivity at the luminance level of
the experiments. 4) Fruit flies may have less contrast adaptation
ability. Or 5) fruit flies may employ an alternate motion detection
mechanism.

Despite the difference between the present results in fruit flies from
those of honey bees at low contrast levels, at high contrasts (above
0.16 Michelson contrast for {\em Drosophila}) flight speed in both
species was regulated around a constant value. This suggests that the
visual system has little trouble estimating self-motion at these
contrast values and that insects regulate flight speed about a set
point using visual information.

{\Large {\sc {\bf (\fig \ref{fig:contrast_speed} near here.) }}}

\section{Conclusions}

The highly automated and real time capabilities of our system allow
unprecedented experimental opportunities. We are currently
investigating the object approach and avoidance phenomenon of fruit
flies illustrated in \fig \ref{fig:analysis}. We are also studying
maneuvering in solitary and competing hummingbirds and the role of
maneuvering in establishing dominance. One of the opportunities made
possible by the molecular biological revolution are powerful new tools
that can be used to visualize and modify the activity of neurons and
neural circuits. By precisely quantifying high level behaviors, such
as the object attraction/repulsion described above, we hope to make
use of these tools to approach the question of how neurons contribute
to the process of behavior.

\appendix
\renewcommand\thesection{Appendix \Alph{section}}
\renewcommand{\theequation}{A-\arabic{equation}}
\setcounter{equation}{0}  

\section{Extended Kalman filter}\label{EKF}

The extended Kalman filter was used as described in Section
\ref{kalman_filtering}. Here, we give the equations using the notation
of this paper. Note that because only the observation process is
non-linear, the process dynamics are specified with (linear) matrix
$\kalmanA$.

The {\em a priori} predictions of these values based on the previous
frame's posterior estimates are
\begin{equation}
\kstateapriori = \kalmanA \kstateapostprev
\end{equation}and
\begin{equation}
\kesterrapriori = \kalmanA \kesterrapostprev \kalmanA^\trnsps + \kalmanQ.
\end{equation}
To incorporate observations, a gain term $\kalmangaintimeless$ is
calculated to weight the innovation arising from the difference
between the {\em a priori} state estimate $\kstateapriori$ and the
observation $\kobsvec$
\begin{equation}
\kalmangain = \frac{ \kesterrapriori \kalmanC^\trnsps } { \kalmanC \kesterrapriori \kalmanC^\trnsps + \kalmanR }.
\end{equation}
The observation matrix, $\kalmanC$, is defined to be the Jacobian of
the observation function (equation \eqnref{eqn:observation_func})
evaluated at the expected state
\begin{equation}
\kalmanC= \left . \frac{ \partial h }{\partial \kstatevec} \right \vert _{\kstateapriori}.
\end{equation}
The posterior estimates are then
\begin{equation}
\kstateapost = \kstateapriori + \kalmangain( \kobsvec_t - \kalmanC \kstateapriori )
\end{equation} and
\begin{equation}
\kesterrapost = (\eyemat - \kalmangain \kalmanC )\kesterrapriori.
\end{equation}

\renewcommand{\theequation}{B-\arabic{equation}}
\setcounter{equation}{0}  

\section{Triangulation}\label{triangulation}

The basic 2D-to-3D calculation finds the best 3D location for two or
more 2D camera views of a point, and is implemented using a linear
least-squares fit of the intersection of $\numcams$ rays defined by
the 2D image points and 3D camera centers of each of the $\numcams$
cameras \citep{hartley_zisserman_2003}.  After correction for radial
distortion, the image of a 3D point on the $i$th camera is
$(u_i,v_i)$. For mathematical reasons, it is convenient to represent
this 2D image point in homogeneous coordinates
\begin{equation} \label{twodeehomogeneous}
\twodeevec_i=(r_i,s_i,t_i),
\end{equation}
such that $u_i = r_i/t_i$ and $v_i=s_i/t_i$. For convenience, we
define the function $\dehomogenize$ to convert from homogeneous to
Cartesian coordinates, thus
\begin{equation} \label{eqn:dehomogenize}
\dehomogenize(\twodeevec)=(u,v)=(r/t,s/t).
\end{equation} The $3\cross4$ camera
calibration matrix $\cammatrix_i$ models the projection from a 3D
homogeneous point $\threedeevec = (X_1, X_2, X_3, X_4)$ (representing
the 3D point with inhomogeneous coordinates $(x,y,z) = (X_1 / X_4,
X_2/X_4, X_3/ X_4)$) into the image point:
\begin{equation} \label{imagepoint}
\twodeevec_i = \cammatrix_i\threedeevec.
\end{equation} By combining the image point
equation \eqnref{imagepoint} from two or more cameras, we can solve
for $\threedeevec$ using the homogeneous linear triangulation method
based on the singular value decomposition as described in
\citet[Sections 12.2 and A5.3]{hartley_zisserman_2003}.

A similar approach can be used for reconstructing the orientation of
the longitudinal axis of an insect or bird. Briefly, a line is fit to
this axis in each 2D image (using $u$, $v$ and $\theta$ from Section
\ref{2D_features}) and, together with the camera center, is used to
represent a plane in 3D space. The best-fit line of intersection of
the $\numcams$ planes is then found with a similar singular value
decomposition algorithm \citep[Section 12.7]{hartley_zisserman_2003}.

\section*{Acknowledgments}
\begin{figure*}
\includegraphics[width=\onefivecolfig]{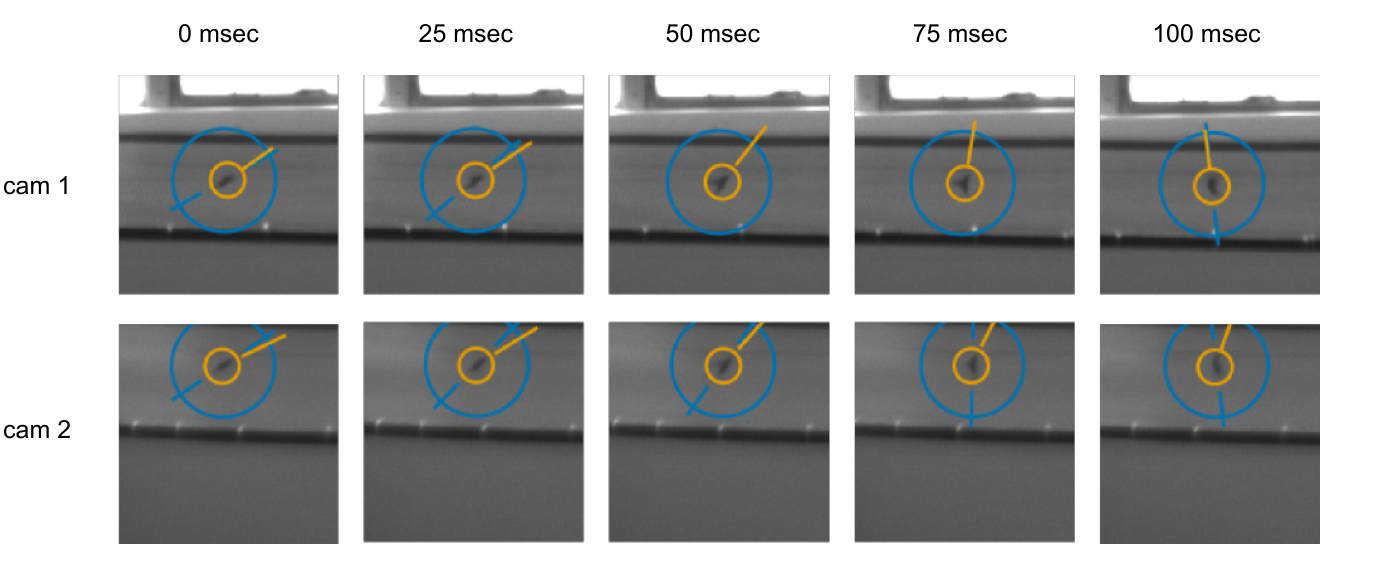}

\caption{Raw images, 2D data extracted from the images, and overlaid
  computed 3D position and body orientation of a hummingbird ({\em
    Calypte anna}). In these images, a blue circle is drawn centered
  on the 2D image coordinates $(\tilde{u},\tilde{v})$. The blue line
  segment is drawn through the detected body axis ($\theta$ in Section
  \ref{data_association}) when eccentricity ($\epsilon$) of the
  detected object exceeds a threshold. The orange circle is drawn
  centered on the 3D estimate of position $(x,y,z)$ reprojected
  through the camera calibration matrix $\cammatrix$, and the orange
  line segment is drawn in the direction of the 3D body orientation
  vector.}
\label{fig:humdra}
\end{figure*}

The data for {\fig} \ref{fig:humdra} were gathered in collaboration
with Douglas Altshuler. Sawyer Fuller helped with the EKF formulation,
provided helpful feedback on the manuscript and, together with Gaby
Maimon, Rosalyn Sayaman, Martin Peek and Aza Raskin, helped with
physical construction of arenas and bug reports on the software.  Pete
Trautmann provided insight on data association, and Pietro Perona
provided helpful suggestions on the manuscript. This work was
supported by grants from the Packard Foundation, AFOSR
(FA9550-06-1-0079), ARO (DAAD 19-03-D-0004) and NIH (R01 DA022777).

\bibliographystyle{apalike}

\bibliography{ads_master}

\end{document}